\documentclass[runningheads]{llncs}
\usepackage{graphicx,subfigure}
\usepackage{amsmath,amssymb} 
\usepackage{color}
\usepackage{cite}
\usepackage{booktabs}
\usepackage{tabularx}
\usepackage{multirow}
\usepackage{algorithm}
\usepackage{bm}
\usepackage{pifont}
\newcommand{\cmark}{\ding{51}}%
\newcommand{\xmark}{\ding{55}}%

\begin{document}
\title{Unsupervised Person Re-identification by \\ Deep Learning Tracklet Association} 

\titlerunning{Unsupervised Person Re-ID by Deep Learning Tracklet Association}
%
\author{Minxian Li\inst{1,2}
	\and
	Xiatian Zhu\inst{3}
	\and
	Shaogang Gong\inst{2}
}
%
\authorrunning{M. Li, X. Zhu, and S. Gong}
%

\institute{$^1$ Nanjing University of Science and Technology\\
		\email{minxianli@njust.edu.cn}	\\
		$^2$ Queen Mary University of London \\
		\email{s.gong@qmul.ac.uk}	\\
		$^3$ Vision Semantics Limited	\\
		\email{eddy@visionsemantics.com}
}
\maketitle              
\begin{abstract}
Most existing person re-identification (re-id) methods rely on 
{\em supervised} model learning on per-camera-pair {\em manually} labelled pairwise training data.
This leads to poor scalability in practical re-id deployment
due to the lack of exhaustive identity labelling of image positive and negative pairs
for every camera pair. 
In this work, we address this problem 
by proposing an unsupervised re-id deep learning approach
capable of incrementally discovering and exploiting the underlying re-id discriminative
information from {\em automatically} generated person tracklet data from videos
in an end-to-end 
model optimisation.
We formulate a 
{\em Tracklet Association Unsupervised Deep Learning} 
(TAUDL) framework characterised by 
jointly learning per-camera (within-camera) tracklet association
(labelling) and cross-camera
tracklet correlation by maximising the discovery of most likely tracklet relationships 
across camera views.
Extensive experiments 
demonstrate the superiority of the proposed TAUDL model
over the state-of-the-art unsupervised and domain adaptation re-id
methods using six person re-id benchmarking datasets.
\keywords{Person Re-Identification; Unsupervised Learning; Tracklet; Surveillance Video.}
\end{abstract}

\section{Introduction}
Person re-identification (re-id) aims to match the underlying identities 
of person bounding box images detected from
non-overlapping camera views \cite{gong2014person}.
In recent years, 
extensive research attention has been attracted
\cite{li2014deepreid,cheng2016person,ahmed2015improved,subramaniam2016deep,xiao2016learning,wang2016joint,li2017person,chen2017beyond,cho2016improving,hermans2017defense,zhang2017deep,li2018harmonious,farenzena2010person}
to address the re-id problem. 
Most existing re-id methods, in particular deep learning models,
adopt the {\em supervised} learning approach. 
These supervised deep models assume the availability of a large number of {\em manually}
labelled {\em cross-view identity (ID) matching image pairs} for each 
camera pair 
in order to induce a feature representation or a
distance metric function optimised just for that camera pair.
This assumption is inherently limited for generalising a re-id model
to many different camera networks therefore cannot scale in practical
deployments\footnote{Exhaustive manual ID labelling of person image pairs 
for every camera-pair is prohibitively 
expensive as there are a quadratic number of camera pairs in a network.}.

It is no surprise then that person re-id by {\em unsupervised} learning has become
a focus in recent research where per-camera pairwise ID labelled
training data is not required in model learning
\cite{wang2014unsupervised,kodirov2015dictionary,lisanti2015person,kodirov2016person,khan2016unsupervised,wang2016towards,ma2017person,ye2017dynamic,liu2017stepwise,zhao2017person}. 
However, all these classical unsupervised learning models are
significantly weaker in re-id performance than the supervised models.
This is because the lack of cross-view pairwise ID labelled data deprives
a model's ability to learn from strong context-aware ID discriminative
information in order to cope with significant visual appearance change
between every camera pair, as defined by a triplet verification loss function.
An alternative approach is to leverage jointly
{(1)} unlabelled data from a target domain which is freely available,
e.g. videos of thousands of people travelling through a camera view everyday in a public scene;
and 
{(2)} pairwise ID labelled datasets from independent source domains
\cite{want2018Transfer,peng2016unsupervised,fan2017unsupervised,yu2017cross,su2016deep}. 
The main idea is to first learn
a ``view-invariant'' representation from ID labelled source data,
then adapt the model to a target domain by using only unlabelled
target data. This approach makes an implicit assumption that
the source and target domains share some common cross-view
characteristics and a view-invariant representation can be estimated, which is not always true. 

In this work, we consider a {\em pure} unsupervised person re-id deep
learning problem. That is, no ID labelled training data is
assumed, neither cross-view 
nor within-view ID labelling.
Although this learning objective is similar to two domain
transfer models \cite{fan2017unsupervised,want2018Transfer}, both
those models do require {\em suitable}, i.e. visually similar to the
target domain, person identity labelled source domain training data.
Specifically, we consider unsupervised re-id model learning 
by jointly optimising unlabelled person tracklet data {\em
  within-camera} view to be more discriminative and {\em cross-camera}
view to be more associative in an end-to-end manner.

Our {\bf contributions} are:
We formulate a novel unsupervised person re-id deep learning method
using person tracklets without the need for
camera pairwise ID labelled training data,
i.e. {\em unsupervised tracklet re-id discriminative learning}.
Specifically, we propose a {\bf Tracklet Association Unsupervised
  Deep Learning} (TAUDL) model with two key innovations: (1) {\em
  Per-Camera Tracklet Discrimination Learning} that 
optimises ``local'' within-camera tracklet label discrimination for
facilitating cross-camera tracklet association given per-camera independently
created tracklet label spaces. 
(2) {\em
  Cross-Camera Tracklet Association Learning} that maximises
``global'' cross-camera tracklet label association. 
This is formulated as to maximise jointly
cross-camera tracklet similarity and
within-camera tracklet dissimilarity in an end-to-end deep learning framework.

Comparative experiments 
show the advantages of TAUDL over the state-of-the-art
unsupervised and domain adaptation person re-id models using six
benchmarks including three multi-shot image based and three video
based re-id datasets:
CUHK03 \cite{li2014deepreid}, 
Market-1501 \cite{zheng2015scalable}, 
DukeMTMC \cite{ristani2016MTMC}, 
iLIDS-VID \cite{wang2014person},
PRID2011 \cite{hirzer2011person},
and MARS \cite{zheng2016mars}.

\section{Related Work}
Most existing re-id models are built by {\em supervised} model learning
on a separate set of per-camera-pair ID labelled training data
\cite{li2014deepreid,cheng2016person,ahmed2015improved,subramaniam2016deep,xiao2016learning,wang2016joint,li2017person,chen2017beyond,jiao2018deep,wang2016person,cho2016improving,chen2017person,chen2018person,zhu2017fast,wang2018person,hermans2017defense,zhang2017deep,li2018harmonious}.
Hence, their scalability and usability is poor for 
real-world re-id deployments
where no such large training sets are available
for every camera pair.
Classical unsupervised learning methods based on hand-crafted features
offer poor re-id performance \cite{farenzena2010person,ma2017person,
  kodirov2015dictionary,kodirov2016person,khan2016unsupervised,ye2017dynamic,liu2017stepwise,
  lisanti2015person,wang2016towards,wang2014unsupervised,zhao2017person}
when compared to the supervised learning based re-id models.
While a balancing trade-off between model scalability and re-id accuracy
can be achieved by semi-supervised learning
\cite{liu2014semi,wang2016towards}, these models still assume
sufficiently large sized cross-view pairwise labelled data for model training. 
More recently, there are some attempts on unsupervised learning of
domain adaptation models
\cite{want2018Transfer,peng2016unsupervised,fan2017unsupervised,yu2017cross,su2016deep}. 
The main idea is to explore knowledge from pairwise labelled data in
``related'' source domains with model adaptation on unlabelled target
domain data.
Whilst these domain adaptation models perform better than the
classical unsupervised learning methods (Table~\ref{tab:img_SOTA} and Table~\ref{tab:vide_SOTA}),
they requires implicitly similar data distributions and viewing
conditions between the labelled source domain and the unlabelled
target domains. This restricts their scalability to arbitrarily
diverse (and unknown) target domains.

In contrast to all these existing unsupervised learning re-id methods,
the proposed tracklet association based method enables unsupervised
re-id deep end-to-end learning from scratch without any assumption on either the 
scene characteristic similarity between source and target domains, or the complexity of
handling identity label space (or lack of) knowledge
transfer in model optimisation.
Instead, our method directly learns to discover the re-id discriminative knowledge
from {\em unsupervised} tracklet label data automatically generated
and annotated from the video data using a common deep learning network architecture.
Moreover, this method does not assume any overlap of person ID classes across camera views,
therefore scalable to any camera networks without any knowledge about
camera space-time topology and/or time-profiling on people cross-view
appearing patterns \cite{loy_ijcv2010}.
Compared to classical unsupervised methods relying on extra hand-crafted features,
our method learns tracklet based re-id discriminative features from an
end-to-end deep learning process. 
To our best knowledge, this is the {\em first} attempt at unsupervised
tracklet association based person re-id deep learning model without
relying on any ID labelled training data (either videos or images).

\begin{figure}[t]
	\centering
	\includegraphics[width=\textwidth]{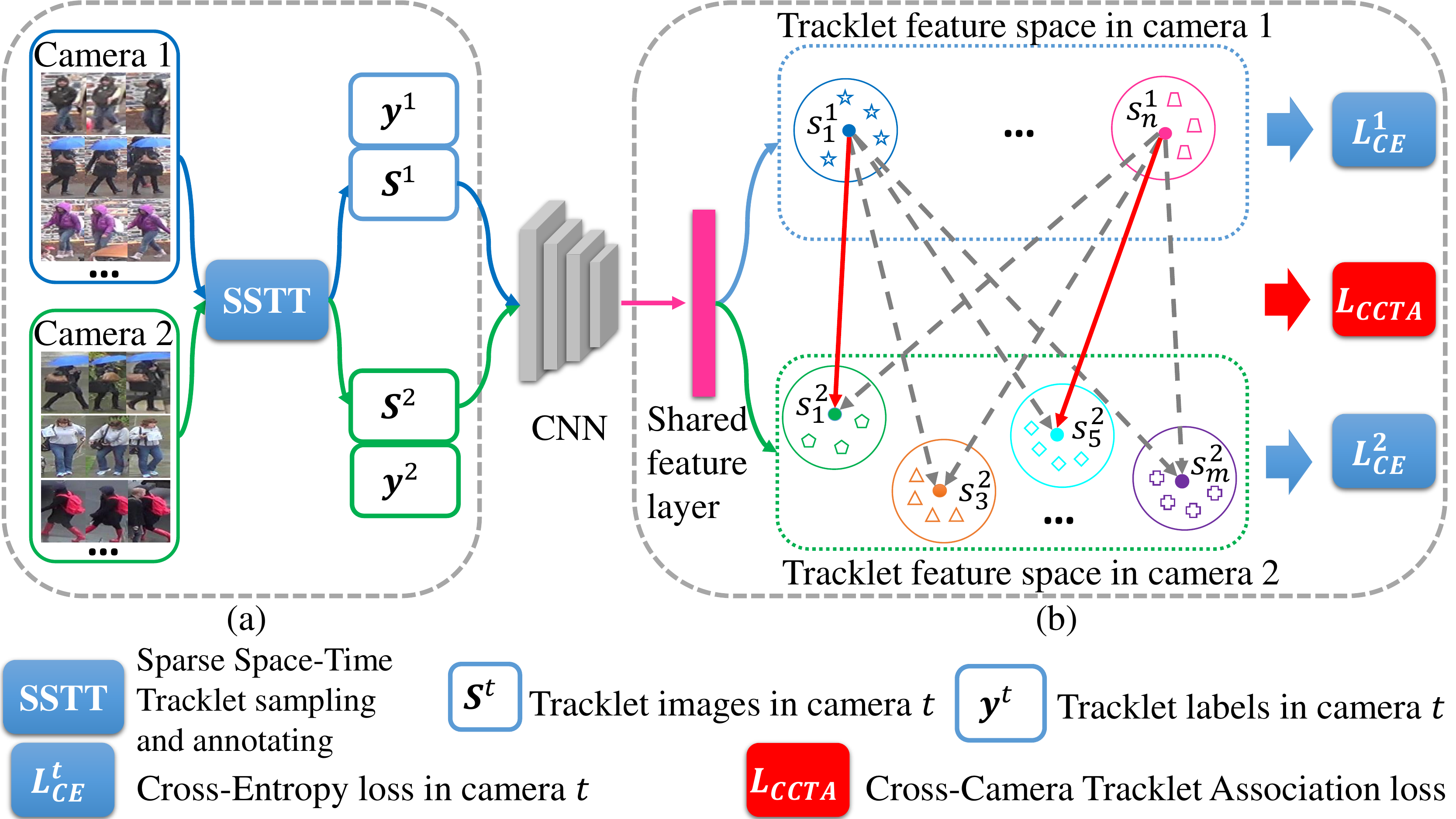}
	\caption{
		An overview of {Tracklet Association Unsupervised Deep Learning} 
		(TAUDL) re-id model: (a) Per-camera unsupervised
                tracklet sampling and label assignment; (b) Joint
                learning of both within-camera tracklet discrimination and
                cross-camera tracklet association in an end-to-end
                global deep learning on tracklets from all the cameras. 
	}
	\label{fig:pipeline}
\end{figure}

\section{Unsupervised Deep Learning Tracklet Association} \label{sec:method}

To overcome the limitation of {\em supervised re-id model training}, 
we propose a novel {\bf Tracklet Association Unsupervised Deep Learning} 
(TAUDL)
approach to person re-id in video (or multi-shot images in general) by
uniquely exploiting person {\em tracklet labelling} obtained by an {\em
  unsupervised} tracklet formation (sampling)
mechanism\footnote{Although object tracklets can be generated by any
  independent single-camera-view multi-object tracking (MOT) models widely
  available today, a conventional MOT model is {\em not} end-to-end optimised for
  cross-camera tracklet association.} {\em without}
any ID labelling of the training data (either cross-view or within-view).
The TAUDL trains a person re-id model in an end-to-end manner in order to
benefit from the inherent overall model optimisation advantages from deep learning.
In the following, we first present a data sampling mechanism for unsupervised
within-camera tracklet labelling (Sec. \ref{sec:tracklet_labelling})
and then describe our model design for cross-camera tracklet association
by joint unsupervised deep learning (Sec. \ref{sec:tracklet_asso_learning}).

\begin{figure}
	\centering
	\subfigure[Temporal sampling]{
		\includegraphics[width=0.6\textwidth]{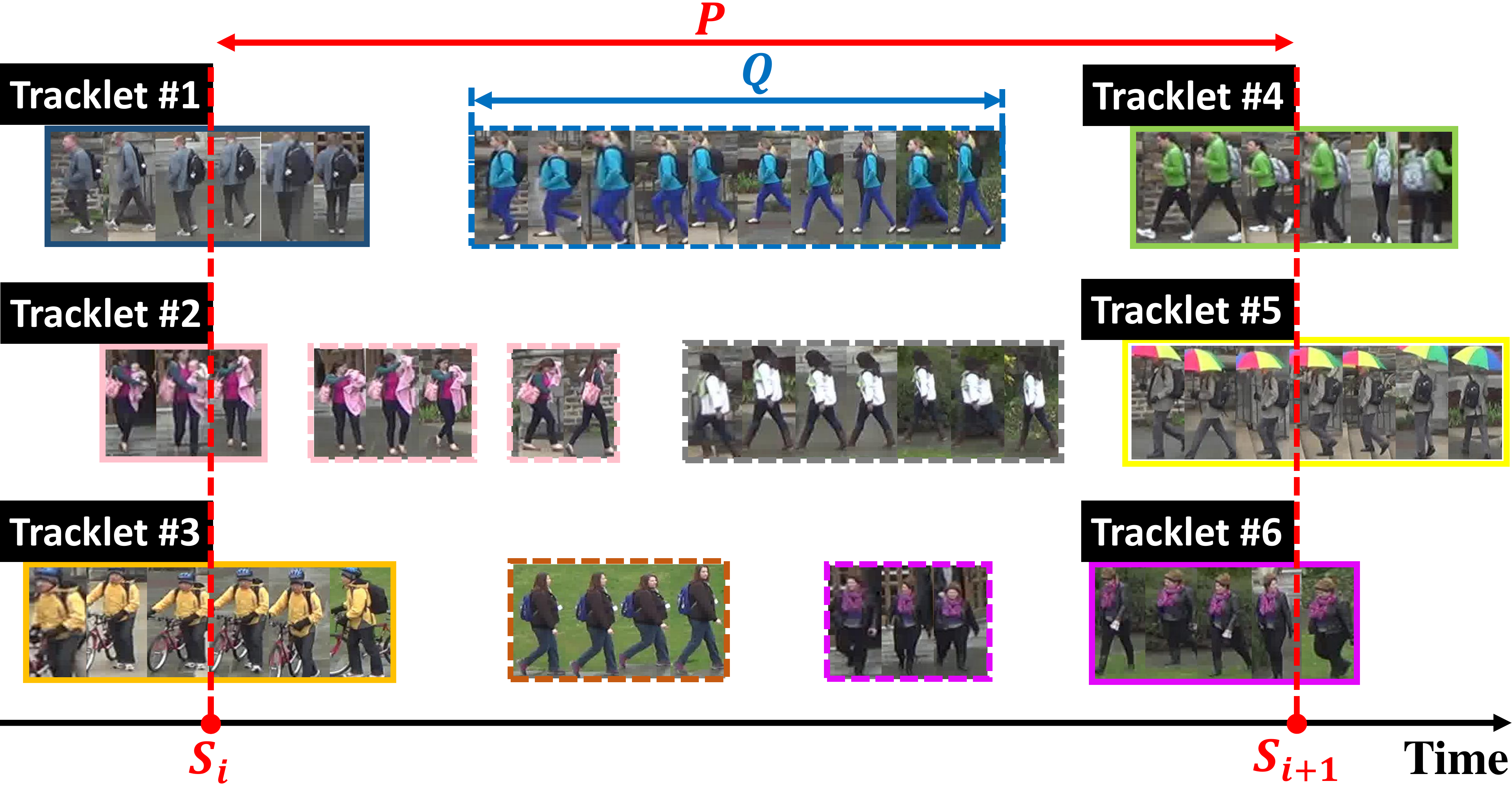}	
	}
	\subfigure[Spatial sampling]{
		\includegraphics[width=0.35\textwidth]{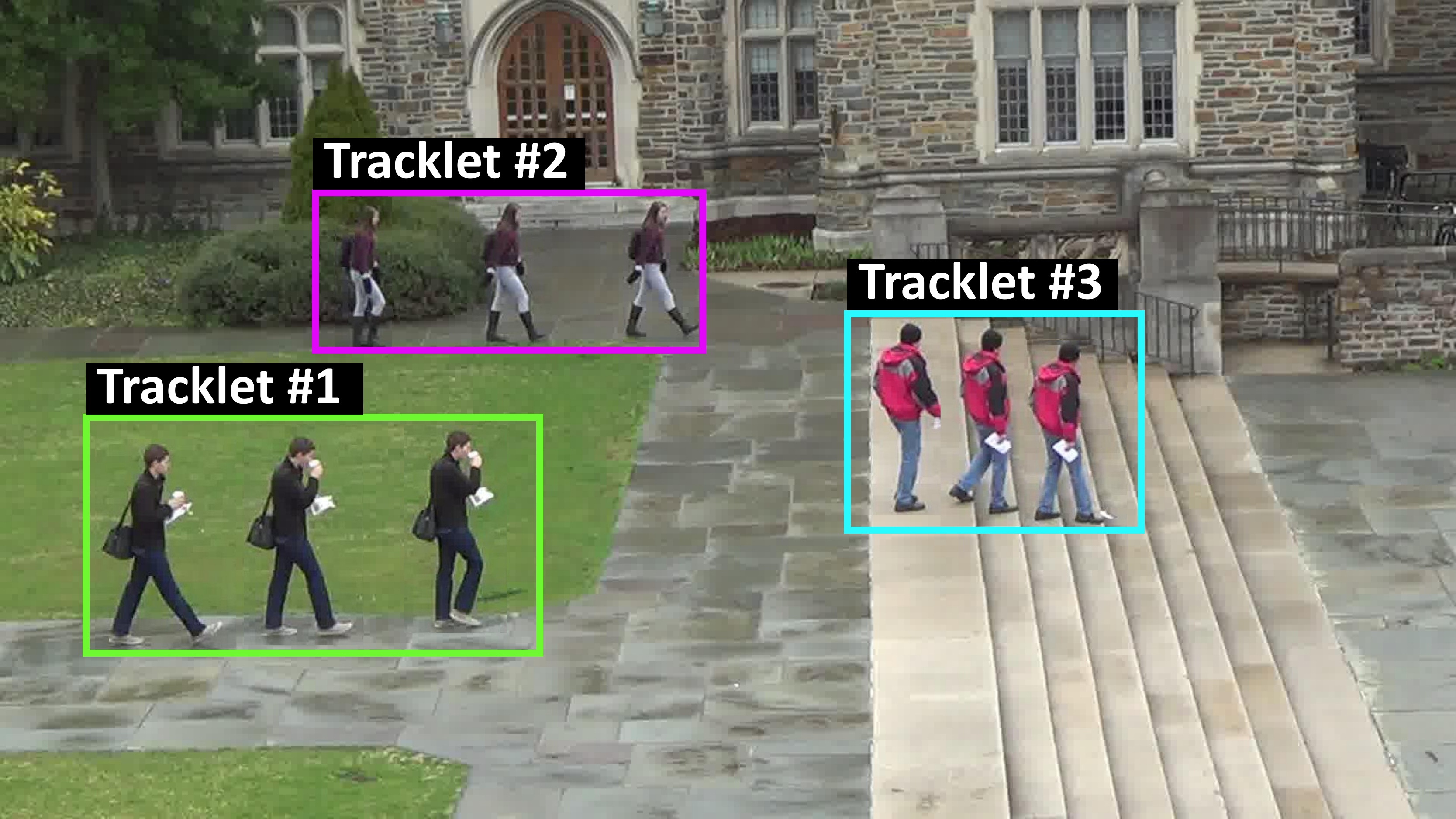}	
	}
	\caption{An illustration of the Sparse Space-Time Tracklet sampling and annotating method
		for unsupervised tracklet labelling.
		Solid box: Sampled tracklets;
		Dashed box: Non-sampled tracklets;
		Each colour represents a distinct person ID.
		(a) Two time instances ($S_i$ and $S_{i+1}$
                indicated by vertical lines) of temporal sampling are shown
		with a time gap $P$ greater than the common transit time $Q$ of a camera view. 
		(b) Three spatially sparse tracklets are formed at a
                given temporal sampling instance.
	}
	\label{fig:tracklet_labelling}
\end{figure}

\subsection{Unsupervised Within-View Tracklet Labelling}
\label{sec:tracklet_labelling}

Given a large quantity of video data from multiple disjoint cameras,
we can readily deploy existing
pedestrian detection and tracking models
\cite{zheng2017unlabeled,ristani2016performance,leal2015motchallenge,zhang2016far}, 
to extract person tracklets. 
In general, the space-time trajectory of a person in a single-camera
view from a public scene is likely to be fragmented into an arbitrary number of short tracklets
due to imperfect tracking and background clutter.
Given a large number of person tracklets per camera, we want to annotate them for deep re-id model learning
in an {\em unsupervised} manner
without any manual identity verification on tracklets.
To this end, we need an automatic tracklet labelling method
to minimise the person ID duplication (i.e. multiple tracklet labels corresponding the
same person ID label) rate among these labelled tracklets.
To this end, 
we propose a {\bf Sparse Space-Time Tracklet} (SSTT) sampling and
label assignment method.

Our SSTT method is built on three observations 
typical in surveillance videos:
{\bf(1)} For most people, 
re-appearing in a camera view is rare during a short time period.
As such, the dominant factor for causing person tracklet duplication
(of the same ID) in
auto-generated person tracklets is trajectory fragmentation, and if we assign every tracklet with a distinct label.
To address this problem, we perform sparse temporal sampling of tracklets
(Fig. \ref{fig:tracklet_labelling}(a)) as follows:
(i) At the $i$-th temporal sampling instance corresponding to a time point $S_i$,
we retrieve all tracklets at time $S_i$
and annotate each tracklet with a distinct label. 
This is based on the factor that
{\bf(2)} people co-occurring at the same time in a single-view but 
at different spatial locations should have distinct ID labels.
(ii) Given a time gap $P$, the next ($(i+1)$-th) temporal sampling and
label assignment is repeated, where $P$ controls the sparsity of the temporal sampling rate.
Based on observation {\bf(3)} that
most people in a public scene travel through a single camera view
in a common time period $Q < P$, 
it is expected that at most one tracklet per person can
be sampled at such a sparse temporal sampling rate (assuming no
re-appearing once out of the same camera view).
Consequently, we can significantly reduce the
ID duplication even in highly crowded scenes with greater degrees of trajectory fragmentation.

To further mitigate the negative effect of inaccurate person detection
and tracking at each temporal sampling instance, we 
further impose a sparse spatial sampling constraint
-- only selecting the co-occurring
tracklets distantly distributed over the scene space (Fig. \ref{fig:tracklet_labelling}(b)).
In doing so, 
the tracklet labels are more likely to be of independent person identities
with minimum ID duplications in each $i$-th temporal sampling instance. 

By deploying this SSTT tracklet labelling method in each camera view, 
we can obtain an independent set of labelled tracklets $\{\bm{S}_i, y_i\}$
per-camera in a camera network,
where each tracklet contains a varying number of 
person bounding boxes as $\bm{S} = \{\bm{I}_1, \bm{I}_2, \cdots \}$. 
Our objective is to use these SSTT labelled tracklets
for optimising a cross-view person re-id deep learning model
without any cross-view ID labelled pairwise training data.

\subsection{Unsupervised Tracklet Association}
\label{sec:tracklet_asso_learning}

Given per-camera independently-labelled tracklets  
$\{\bm{S}_i, y_i\}$ generated by SSTT, 
we perform {\em tracklet label re-id discriminative learning}
without person ID labels in a 
conventional classification deep learning framework.
To that end, we formulate a {\bf Tracklet Association Unsupervised Deep Learning} (TAUDL) model.
The overall design of our TAUDL architecture is shown
in Fig. \ref{fig:pipeline}.
The TAUDL contains two model components:
{\bf(I)} {\em Per-Camera Tracklet Discrimination Learning} 
with the aim to optimise ``local'' (within-camera) tracklet label discrimination
for facilitating cross-camera tracklet association
given independently created tracklet label spaces in different camera views.
{\bf(II)} {\em Cross-Camera Tracklet Association Learning}
with the aim to maximise ``global'' (cross-camera) tracklet label association.
The two components integrate as a whole in a single deep learning 
network architecture, learn jointly and mutually benefit each other
in an incremental end-to-end manner.

\begin{figure} [h]
	\centering
	\subfigure{
		\includegraphics[width=0.46\textwidth]{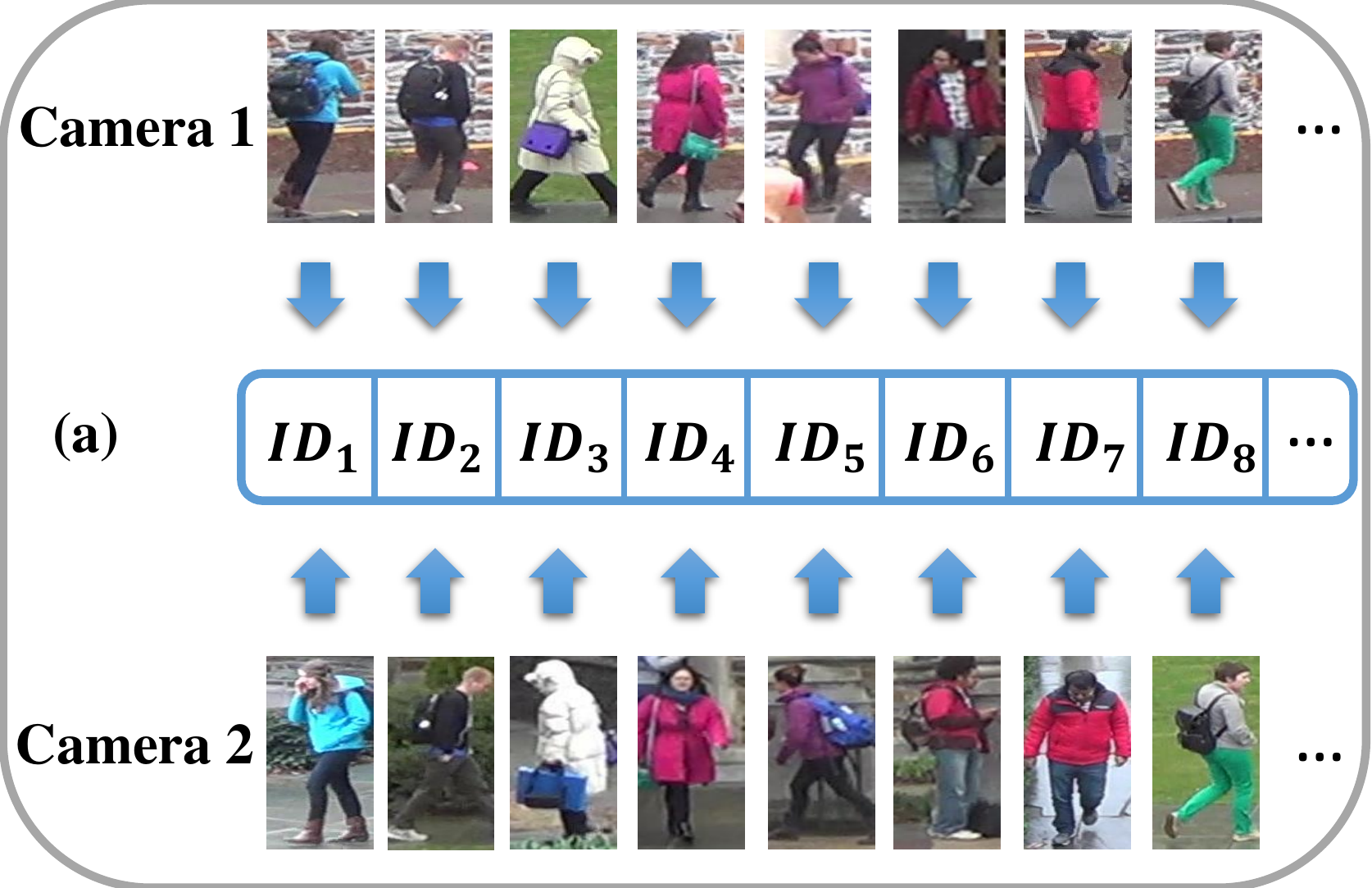}	
	}
	\subfigure{
		\includegraphics[width=0.46\textwidth]{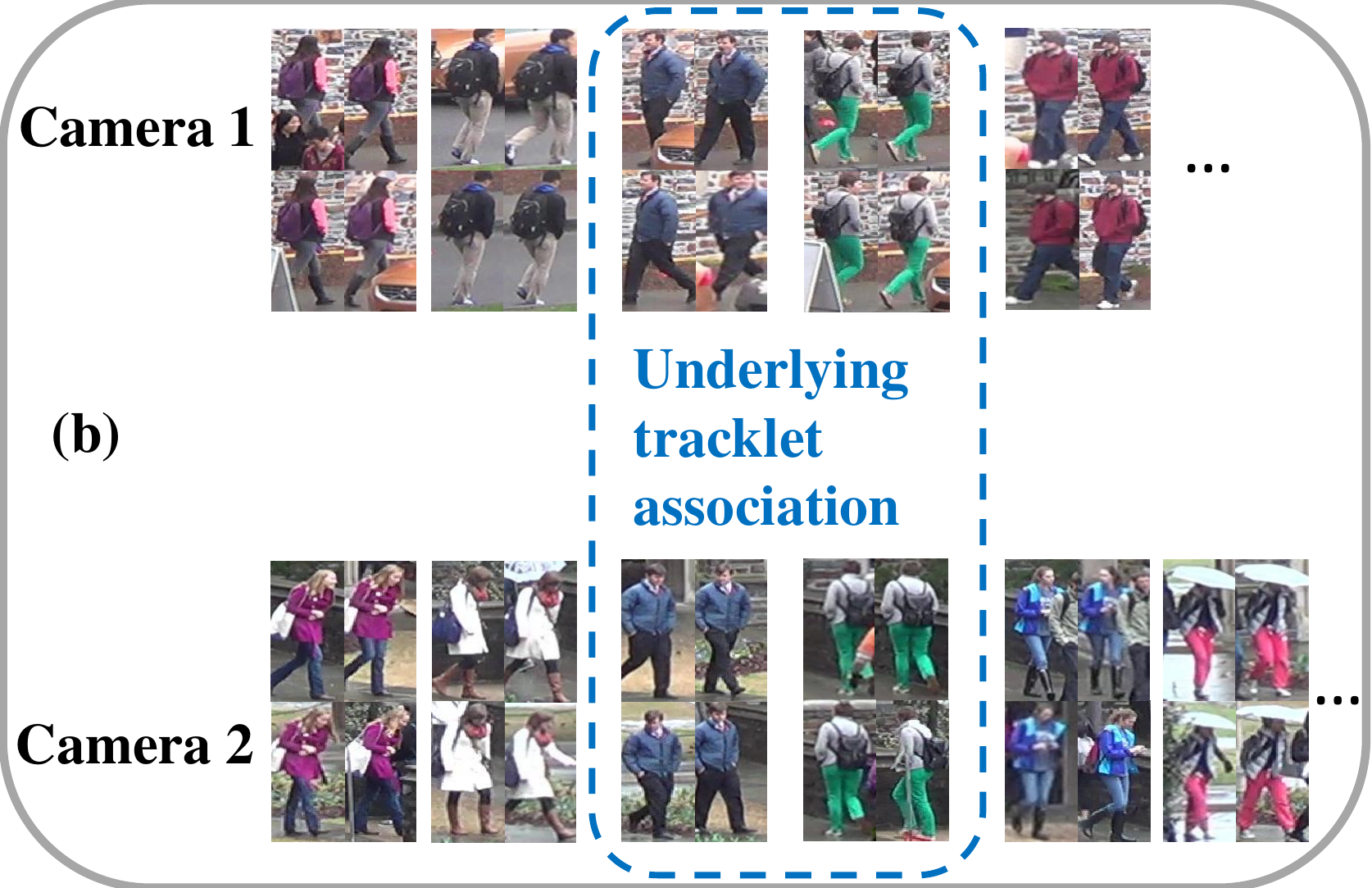}	
	}
	\caption{Comparing
		(a) Fine-grained {\em explicit instance-level}
          cross-view ID labelled image pairs for supervised person
          re-id model learning and 
		(b) Coarse-grained {\em latent group-level} cross-view
                tracklet (a multi-shot group) label correlation for ID label-free
                (unsupervised) person re-id learning using TAUDL. 
	}
	\label{fig:label_info}
\end{figure}

\noindent{\bf (I) Per-Camera Tracklet Discrimination Learning }
For accurate cross-camera tracklet association, 
it is important to formulate a robust image feature representation for
describing the person appearance of each tracklet
that helps cross-view 
person re-id association.
However, it is sub-optimal to achieve ``local'' per-camera tracklet
discriminative learning using only per-camera independent tracklet labels
without ``global'' cross-camera tracklet correlations. We wish to
optimise jointly both local tracklet within-view discrimination and global
tracklet cross-view association.
To that end, we design a Per-Camera Tracklet Discrimination (PCTD) learning algorithm.
Our key idea is that, instead of relying on 
the conventional fine-grained {\em explicit instance-level} cross-view
ID pairwise supervised learning
(Fig. \ref{fig:label_info}(a)),
we learn to maximise coarse-grained {\em latent group-level}
cross-camera tracklet association by set correlation
(Fig. \ref{fig:label_info}(b)).

Specifically, we treat each individual camera view separately
by optimising per-camera labelled tracklet discrimination as a
classification task against the tracklet labels per-camera 
(not person ID labels). 
Therefore, we have a total of $T$ different tracklet classification tasks
each corresponding to a specific camera view.
Importantly, we further formulate these $T$ classification tasks
in a multi-branch architecture design where 
every task shares the {\em same} feature representation 
whilst enjoys an individual classification branch (Fig. \ref{fig:pipeline}(b)).
Conceptually, this model design is in a spirit of the multi-task learning principle \cite{evgeniou2004regularized,ando2005framework}.

Formally, given unsupervised training data $\{\bm{I}, y\}$ 
extracted from a camera view $t \in \{1, \cdots,T\}$,
where $\bm{I}$ specifies a tracklet frame and $y \in \{1,\cdots,M_t\}$
the tracklet label (obtained as in Sec. \ref{sec:tracklet_labelling})
with a total of $M_t$ different labels,
we adopt the softmax Cross-Entropy (CE) loss function
to optimise the corresponding classification task (the $t$-th branch).
The CE loss on a training image sample $(\bm{I}, y)$ is computed as: 
\begin{equation}
\mathcal{L}_\text{ce}=
{-}{\log}\Big(\frac{\exp({\bm{W}_{y}^{\top} {\bm x}})}
{\sum_{k=1}^{M_t}\exp({\bm{W}_{k}^{\top} {\bm x}})}\Big), 
\label{eq:CE_loss}
\end{equation}
where $\bm{x}$ specifies the feature vector of $\bm{I}$
extracted by the task-shared feature representation component 
and
$\bm{W}_{y}$ the $y$-th class prediction function parameters. 
Given a mini-batch, 
we compute the CE loss for each such training sample
w.r.t. the respective tracklet label space
and utilise their average to form the model learning supervision as:
\begin{equation}
\mathcal{L}_\text{pctd}= \frac{1}{N_\text{bs}} \sum_{t=1}^{T}
\mathcal{L}_\text{ce}^t, 
\label{eq:PCTD_loss}
\end{equation}
where $\mathcal{L}_\text{ce}^t$ denotes the CE loss summation of training samples
from the $t$-th camera among a total of $T$ and 
$N_\text{bs}$ the batch size.

\textbf{\em Discussion}: 
In PCTD, the deep learning objective loss function
(Eqn. \eqref{eq:CE_loss}) aims to optimise by supervised learning
person tracklet discrimination {\em within} each camera view without
any knowledge on {\em cross-camera} tracklet association. 
However, when jointly learning all the per-camera tracklet
discrimination tasks together, the learned representation model is
somewhat {\em implicitly} and {\em collectively} cross-view
tracklet discriminative in a latent manner, due to the existence of cross-camera tracklet
correlation.
In other words,
the shared feature representation is optimised {\em concurrently} to
be discriminative for tracklet discrimination in multiple camera
views, therefore propagating model discriminative learning from per-camera
to cross-camera. We will evaluate the effect of this model design in
our experiments (Table \ref{tab:PCJL}).

\noindent{\bf (II) Cross-Camera Tracklet Association Learning }
While the PCTD algorithm described above 
achieves somewhat global (all the camera views) tracklet discrimination implicitly, the learned
model representation remains sub-optimal due to the lack of {\em
  explicitly} optimising cross-camera tracklet association at the
fine-grained instance level.
It is significantly harder to impose cross-view person re-id
discriminative model learning without camera pairwise ID labels.
To address this problem,
we introduce a Cross-Camera Tracklet Association (CCTA) loss function.
The CCTA loss is formulated based on the idea of 
{\em batch-wise incrementally aligning cross-view per tracklet feature
  distribution} in the shared multi-task learning feature
space. 
Critically, CCTA integrates seamlessly with PCTD to jointly optimise
model learning on discovering cross-camera tracklet association for
person re-id in a single end-to-end batch-wise learning process.

Formally, given a mini-batch including a subset of tracklets
$\{(\bm{S}_i^t, y_i^t)\}$ where $\bm{S}_i^t$ specifies
the $i$-th tracklet from $t$-th camera view with
the label $y_i^t$ where tracklets in a mini-batch come from all the
camera views,
we want to establish for each in-batch tracklet a discriminative
association with other tracklets from different camera views.
In absence of person identity pairwise labelling as a learning constraint,
we propose to align {\em similar} and {\em dissimilar} tracklets in each
mini-batch given the up-to-date shared multi-task (multi-camera)
feature representation from optimising PCTD.
More specifically,
for each tracklet $\bm{S}_i^t$, 
we first retrieve $K$ cross-view nearest tracklets $\mathcal{N}_i^t$
in the feature space,
with the remaining $\tilde{\mathcal{N}}_i^t$
considered as dissimilar ones.
We then impose a soft discriminative structure constraint
by encouraging the model to pull $\mathcal{N}_i^t$
close to $\bm{S}_i^t$ whilst 
to push away $\tilde{\mathcal{N}}_i^t$ from $\bm{S}_i^t$.
Conceptually, this is a per-tracklet cross-view
data structure distribution alignment.
To achieve this, 
we formulate a CCTA deep learning objective loss for each tracklet $\bm{S}_i^t$ in a training mini-batch as:
\begin{equation}\label{eq:CCTA}
\mathcal{L}_\text{ccta}
=
-\log \frac
{\sum_{\bm{z}_k\in \mathcal{N}_i^t}\exp({-\frac{1}{2\sigma^2}\parallel \bm{s}^{t}_{i} - \bm{z}_k \parallel_{2}})}
{\sum_{t'=1}^T \sum_{j=1}^{n_j} \exp({-\frac{1}{2\sigma^2}\parallel \bm{s}^{t}_{i} - \bm{s}_{j}^{t'} \parallel_{2}})}, 
\end{equation}
where $n_j$ denotes the number of in-batch tracklets from $j$-th camera view,
$T$ the camera view number, 
$\sigma$ a scaling parameter, 
$\bm{s}_i^t$ the up-to-date feature representation of the tracklet $\bm{S}_i^t$.
Given the incremental iterative deep learning nature,
we represent a tracklet $\bm{S}$ by
the average of its in-batch frames' feature vectors
on-the-fly.
Hence, the tracklet representation is kept up-to-date 
without the need for maintaining external per-tracklet feature representations.

\textbf{\em Discussion}:
The proposed CCTA loss formulation is conceptually 
similar to the Histogram Loss \cite{Ustinova2016hist}
in terms of distribution alignment.
However, the Histogram Loss is a {\em supervised} loss that requires
supervised label training data,
whilst the CCTA is purely {\em unsupervised} 
and derived directly from feature similarity measures. 
CCTA is also related to the surrogate (artificially built)
class based unsupervised deep learning loss formulations
\cite{bautista2017deep,bautista2016cliquecnn},
by not requiring groundtruth class-labelled data in model training.
Unlike CCTA without the need for creating surrogate classes, 
the surrogate based models not only require additional global data clustering,
but also are sensitive to the clustering quality and initial feature selection.
Moreover, they do not consider the label distribution alignment 
across cameras and label spaces for which
the CCTA loss is designed.

\noindent {\bf Joint Loss Function }
After merging the CCTA and PCTD learning constraints, we obtain the final 
model objective function as:
\begin{equation}\label{eq:loss_TAUDL}
\mathcal{L}_\text{taudl}= 
(1-\lambda)\mathcal{L}_\text{pctd} + \lambda \mathcal{L}_\text{ccta}, 
\end{equation}
where $\lambda$ is a weighting parameter estimated by cross-validation.
Note that $\mathcal{L}_\text{pctd}$ is an average loss term at the
tracklet individual image level
whilst $\mathcal{L}_\text{ccta}$ at the tracklet group (set) level,
both derived from the same training batch concurrently.
As such, the overall TAUDL method naturally enables end-to-end deep model learning
using the Stochastic Gradient Descent optimisation algorithm.

\section{Experiments}

\noindent \text{\bf Datasets }
To evaluate the proposed TAUDL model, we tested both
video ({MARS} \cite{zheng2016mars}, iLIDS-Video \cite{wang2014person}, PRID2011 \cite{hirzer2011person})
and 
image (CUHK03 \cite{li2014deepreid}, Market-1501
\cite{zheng2015scalable}, DukeMTMC \cite{ristani2016MTMC,zheng2017unlabeled})
based person re-id benchmarking datasets. 
In previous studies, these datasets were mostly evaluated separately.
We consider since recent large sized image based re-id datasets were typically constructed 
by sampling person bounding boxes from video, these image datasets share
similar characteristics of those video based datasets.
We adopted the standard person re-id setting on training/test ID split
and the test protocols (Table \ref{tab:dataset_stats}).

\begin{figure}[t]
\centering
\includegraphics[width=0.15\textwidth]{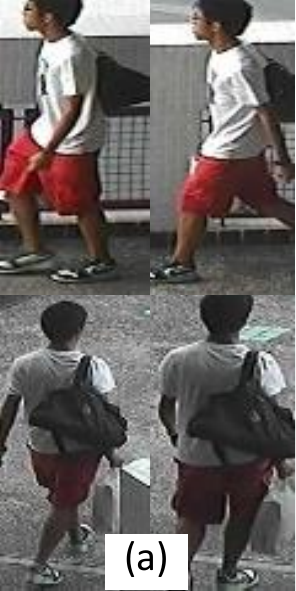}
\includegraphics[width=0.15\textwidth]{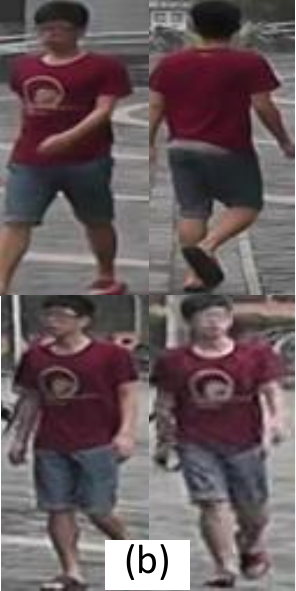}
\includegraphics[width=0.15\textwidth]{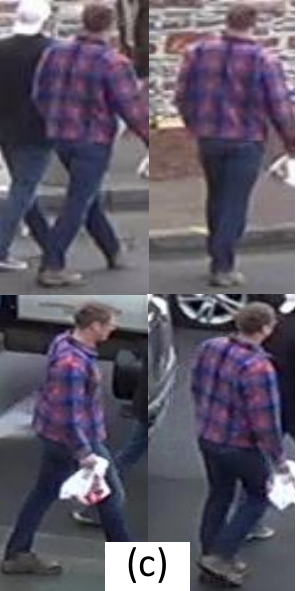}
\includegraphics[width=0.15\textwidth]{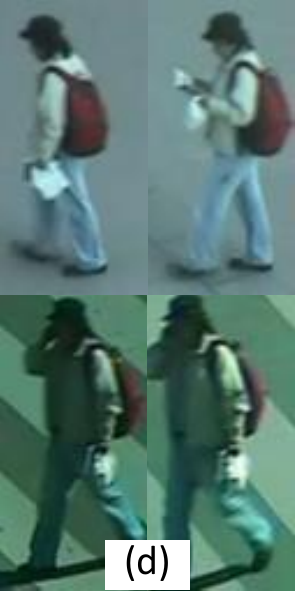}
\includegraphics[width=0.15\textwidth]{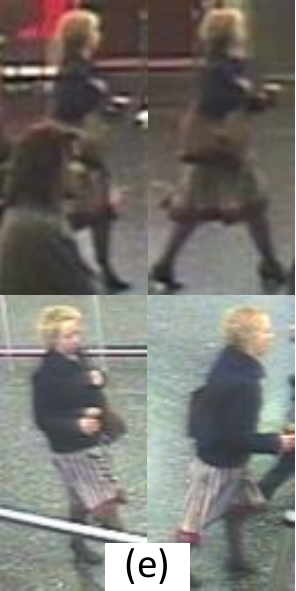}
\includegraphics[width=0.15\textwidth]{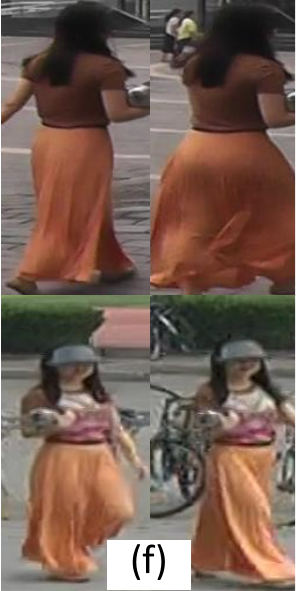}
\caption{Example cross-view matched image/tracklet pairs from 
	(a) CUHK03, (b) Market-1501, (c) DukeMTMC, (d) PRID2011, (e) iLIDS-VID, (f) MARS.
}
\label{fig:dataset_img}
\end{figure}

\begin{table}[h] 
	\centering
	\setlength{\tabcolsep}{0.25cm}
	\caption{
		Dataset statistics and
		evaluation setting.
	}
	\begin{tabular}{c||c|c|c|c|c}
		\hline 
		Dataset  & 
		{\# ID} & 
		{\# Train } & 
		{\# Test} &
		{\# Images} & 
		{\# Tracklet} \\ \hline \hline
		iLIDS-VID \cite{wang2014person}
		& 300 & 150 & 150 & 43,800 & 600 \\
		PRID2011 \cite{hirzer2011person}
		& 178 & 89 & 89 & 38,466 & 354 \\
		MARS \cite{zheng2016mars} 
		& 1,261 & 625 & 636 & 1,191,003 & 20,478 \\ 
		\hline
		CUHK03 \cite{li2014deepreid}
		& 1,467 & 767 & 700 & 14,097 & 0 \\
		Market-1501 \cite{zheng2015scalable}
		& 1,501& 751 & 750  & 32,668 & 0 \\
		DukeMTMC \cite{ristani2016MTMC}
		& 1,812 & 702 & 1,110 & 36,411 & 0\\
		\hline
	\end{tabular}
	\label{tab:dataset_stats}
\end{table}

\noindent \textbf{Tracklet Label Assignment }
For all six datasets, we cannot perform real SSTT tracklet sampling
and label assignment due to
no information available on spatial and temporal location w.r.t. the original video data.
In our experiment, we instead conducted simulated SSTT to 
obtain the per-camera tracklet/image labels.  
For all datasets, we assume no re-appearing subjects per camera (very
rare in these datasets)
and sparse spatial sampling.
As both iLIDS-VID and PRID2011
provide only one tracklet per ID per camera (i.e. no fragmentation),
it is impossible to have per-camera ID duplication.
Therefore, each tracklet is assigned a unique label.
The MARS gives multiple tracklets per ID per camera. 
Based on SSTT,
at most only one tracklet can be sampled for each ID per camera
(see Sec. \ref{sec:tracklet_labelling}).
Therefore, a MARS tracklet per ID per camera was randomly selected and
assigned a label.
For all {image based datasets}, 
we assume all images per ID per camera were drawn from 
a single tracklet, same as in iLIDS-VID and PRID2011.
The same tracklet label assignment procedure was adopted as above.

\noindent \textbf{Performance Metrics } 
We use the common cumulative matching characteristic (CMC) and mean Average Precision (mAP) metrics \cite{zheng2015scalable}.

\noindent \textbf{Implementation Details }
We adopted an ImageNet pre-trained ResNet-50 \cite{he2016deep} as the backbone
in evaluating the proposed TAUDL method.
We set the feature dimension of the camera-shared representation space derived on top of ResNet-50
to 2,048.
Each camera-specific branch contains one FC classification layer.
Person images are resized to $256\! \times \! 128$ for all datasets.
To ensure that each batch has the capacity of containing person images from all cameras, 
we set the batch size to 384 for all datasets.
For balancing the model learning speed over different cameras,
we randomly selected the same number of training frame images
per camera when sampling each mini-batch. 
We adopted the Adam optimiser \cite{kingma2014adam} with the initial learning rate of $3.5 \!\times\! 10^{-4}$. 
We empirically set $\lambda\!=\!0.7$ for Eq.~\eqref{eq:loss_TAUDL},
$\sigma\!=\!2$ for Eq.~\eqref{eq:CCTA}, and $K\!=\!T/2$ ($T$ is the number of cameras) for cross-view nearest tracklets $\mathcal{N}_i^t$ in Eq.~\eqref{eq:CCTA} for all the experiments.

\subsection{Comparisons to State-Of-The-Arts}

We compared two different sets of state-of-the-art methods
on image and video re-id datasets, due to the independent studies on them 
in the literature.

\begin{table}[h]
	\centering
	\setlength{\tabcolsep}{0.2cm}
	\caption{Unsupervised re-id on image datasets. 
		$1^\text{st}$/$2^\text{nd}$ best results are in \textbf{\color{red}red}/\textbf{\color{blue}blue}.}
	\label{tab:img_SOTA}
	\begin{tabular}
		{c||c|c||c|c||c|c}
		\hline
		{Dataset}						
		& \multicolumn{2}{c||}{CUHK03\cite{li2014deepreid}}      
		& \multicolumn{2}{c||}{Market-1501\cite{zheng2015scalable}}                    	
		& \multicolumn{2}{c}{DukeMTMC\cite{zheng2017unlabeled}}    						\\ \hline 
		{Metric(\%)}		& Rank-1	& mAP		& Rank-1	& mAP 		& Rank-1	& mAP
		\\ \hline \hline
		Dic\cite{kodirov2015dictionary}	& 36.5	& -         & 50.2	& 22.7		& - 		& -
		\\
		ISR\cite{lisanti2015person}		& 38.5   & -         & 40.3 	& 14.3 		& -         & - 
		\\ 
		RKSL\cite{wang2016towards}		& 34.8   & -         & 34.0   & 11.0          & -         & - 
		\\ \hline
		SAE\cite{lee2008sparse}			& 30.5   & -         & 42.4   & 16.2          & -         & -
		\\
		JSTL\cite{xiao2016learning}		& 33.2   & -         & 44.7   & 18.4          & -       	& - 
		\\
		AML\cite{ye2007adaptive}			& 31.4	& -         & 44.7   & 18.4		& -        & - 
		\\
		UsNCA\cite{qin2015unsupervised}	& 29.6	& -         & 45.2   & 18.9		& -        & - 
		\\
		CAMEL \cite{yu2017cross}		
		& \textbf{\color{blue}39.4} 		& -       
		& {54.5} 					& {26.3} 
		& -                             	 	& -
		\\
		PUL \cite{fan2017unsupervised}	
		& -                     & -                             
		& 44.7                & 20.1           
		& 30.4			& 16.4        	    	
		\\ 
		TJ-AIDL \cite{want2018Transfer}
		& - 				& - 
		&\bf \color{blue} 58.2 	&\bf\color{blue} 26.5 
		&\bf\color{blue} 44.3 		&\bf\color{blue} 23.0 \\
		\hline
		\textbf{TAUDL}                                	
		& \textbf{\color{red}44.7} 	& \textbf{\color{red}31.2} 
		& \textbf{\color{red}63.7}	& \textbf{\color{red}41.2} 
		& \textbf{\color{red}61.7} 	& \textbf{\color{red}43.5}
		\\ \hline
		GCS \cite{chen2018group}({\em Supervised})	& 88.8	& 97.2         & 93.5	& 81.6		& 84.9 	& 69.5
		\\ \hline
	\end{tabular}
\end{table}

\noindent {\bf Unsupervised Person Re-ID on Image Datasets }
Table \ref{tab:img_SOTA}
shows the unsupervised re-id performance
of the proposed TAUDL and 10 state-of-the-art methods including 
3 hand-crafted feature based methods 
(Dic \cite{kodirov2015dictionary}, ISR \cite{lisanti2015person}, RKSL \cite{wang2016towards}) and 
7 auxiliary knowledge (identity/attribute) transfer based models 
(AE \cite{lee2008sparse},
AML \cite{ye2007adaptive}, 
UsNCA \cite{qin2015unsupervised}, 
CAMEL \cite{yu2017cross},
JSTL \cite{xiao2016learning},
PUL\cite{fan2017unsupervised},
TJ-AIDL \cite{want2018Transfer}).
These results show:
{\bf(1)}
Among existing methods,
the knowledge transfer based method is superior,
e.g. on CUHK03, Rank-1 39.4\% by CAMEL vs. 36.5\% by Dic; On
Market-1501, 58.2\% by TJ-AIDL vs. 50.2\% by Dic.
To that end, CAMEL benefits from learning on $7$ different
person re-id datasets of diverse domains
(CUHK03\cite{li2014deepreid}, CUHK01\cite{li2012human}, PRID \cite{hirzer2011person}, VIPeR \cite{gray2008viewpoint}, 3DPeS\cite{baltieri20113dpes}, i-LIDS\cite{prosser2010person}, Shinpuhkan\cite{kawanishi2014shinpuhkan2014}) 
including a total of 44,685 images and 3,791 identities;
TJ-AIDL utilises labelled Market-1501 (750 IDs and 27 attribute classes) or DukeMTMC (702 IDs and 23 attribute classes) as source training data.
{\bf (2)} Our new model TAUDL outperforms all competitors with significant margins.
For example, the Rank-1 margin by TAUDL over TJ-AIDL
is 5.5\% (63.7-58.2) on Market-1501 and 17.4\% (61.7-44.3) on DukeMTMC.
Moreover, it is worth pointing out that TAUDL dose not benefit from any
additional labelled source domain training data as compared to TJ-AIDL.
TAUDL is potentially more scalable due to no need to consider source
and target domains similarities. 
{\bf (3)} 
Our TAUDL is simpler to train
with a simple end-to-end model learning, as compared to the alternated
deep CNN training and clustering required by PUL and a two-stage model
training of TJ-AIDL. 
These results show both the performance advantage and 
model design superiority of the proposed TAUDL model
over a wide variety of state-of-the-art re-id models.

\begin{table}[h]
	\centering
	\setlength{\tabcolsep}{0.07cm}
	\caption{Unsupervised re-id on video datasets.
		$1^\text{st}$/$2^\text{nd}$ best results are in \textbf{\color{red}red}/\textbf{\color{blue}blue}.
	}
	\label{tab:vide_SOTA}
	\begin{tabular}
		{c||c|c|c||c|c|c||c|c|c|c}
		\hline
		Dataset		    						
		& \multicolumn{3}{c||}{PRID2011 \cite{hirzer2011person}}      	
		& \multicolumn{3}{c||}{iLIDS-VID \cite{wang2014person}}		
		& \multicolumn{4}{c}{MARS \cite{zheng2016mars}} 										
		\\ \hline\hline
		Metric(\%) 							
		& R1 	& R5 	& R20 				
		& R1 	& R5 	& R20 		 		
		& R1 	& R5	& R20 			& mAP
		\\ \hline \hline
		DTW \cite{ma2017person}            		
		& 41.7   & 67.1   & 90.1		& 31.5   & 62.1   & 82.4	    	 & -      	& - 		& -       	& -
		\\
		GRDL \cite{kodirov2016person}		
		& 41.6	& 76.4	& 89.9				& 25.7	& 49.9	& 77.6				& 19.3			& 33.2			& 46.5			& 9.56			\\
		UnKISS \cite{khan2016unsupervised}	
		& 58.1	& 81.9	& 96.0				
		& 35.9	&\color{blue} \bf 63.3	& \color{red} \bf 83.4				
		& 22.3	& 37.4	& 53.6	& 10.6
		\\ 
		SMP \cite{liu2017stepwise}            		
		& \color{red} \bf 80.9   & \color{red} \bf 95.6   & \color{red} \bf99.4				
		& \color{red} \bf 41.7   & \color{red} \bf 66.3   & 80.7	    	 		
		& 23.9   				  & 35.8				    & 44.9			& 10.5
		\\ 
		DGM+MLAPG \cite{ye2017dynamic}     
		&\color{blue} \bf  73.1   				  
		&\color{blue} \bf  92.5   	
		&\color{blue} \bf  99.0
		&\color{blue} \bf 37.1   & 61.3  	& 82.0
		& 24.6   				  & 42.6		& 57.2			& 11.8 			\\ 
		DGM+IDE \cite{ye2017dynamic}         	
		&56.4   & 81.3  &96.4
		& 36.2   				  & 62.8   			    & \color{blue} \bf82.7
		& \textbf{\color{blue}{36.8}}   & \textbf{\color{blue}{54.0}}   & \textbf{\color{blue}{68.5}}   & \textbf{\color{blue}{21.3}}
		\\ 
		\hline
		\textbf{TAUDL} 
		& 49.4 	& 78.7	& 98.9
		& 26.7   & 51.3	& 82.0 
		& \textbf{\color{red}43.8}	& \textbf{\color{red}59.9}	& \textbf{\color{red}72.8}	& \textbf{\color{red}29.1} 
		\\ \hline
		{QAN \cite{liu2017quality}({\em Supervised})}
		& 90.3	& 98.2   & 100.0		& 68.0	& 86.8 	& 97.4 		& 73.7	& 84.9 	& 91.6 	& 51.7
		\\ \hline
	\end{tabular}
\end{table}

\noindent {\bf Unsupervised Person Re-ID on Video Datasets }
We compared the proposed TAUDL with six 
state-of-the-art unsupervised video person re-id models.
Unlike TAUDL, all these existing models are not end-to-end deep learning methods
with either hand-crafted or separately trained deep features as model input.
Table \ref{tab:vide_SOTA} shows
that TAUDL outperforms all existing video-based person re-id models on
the large scale video dataset MARS,
e.g. by a Rank-1 margin of 7.0\% (43.8-36.8) over
the best competitor DGM+IDE (which additionally using the ID label information
of one camera view for model initialisation). However, TAUDL is
inferior than some of the existing models on the two small benchmarks
iLIDS-VID (300 training tracklets) and PRID2011 (178 training tracklets), 
in comparison to its performance on the MARS benchmark (8,298 training tracklets).
This shows that TAUDL does need sufficient tracklet data from larger
video datasets in order to have its performance advantage. As the
tracklet data required are not manually labelled, this requirement is not a
hindrance to its scalability to large scale data. 
Quite the contrary,
TAUDL works the best when large scale unlabelled video data is available.
A model would benefit particularly from pre-training using TAUDL on large auxiliary unlabelled video
data from similar camera viewing conditions.

\subsection{Component Analysis and Discussions}

\textbf{Effectiveness of Per-Camera Tracklet Discrimination } The PCTD component
was evaluated by comparing 
a baseline that treats all cameras together 
by concatenating per-camera tracklet label sets
and deploying the Cross-Entropy loss
to learn a unified classification task.
We call this baseline Joint-Camera Classification (JCC).
In this analysis, we do not consider the cross-camera tracklet association component
for a clear evaluation.
Table \ref{tab:PCJL} shows that 
our PCTD design is significantly superior
over the JCC learning algorithm,
e.g. achieving Rank-1 gain of 
4.0\%, 34.6\%, 36.3\%, and 19.9\% on
CUHK03, Market-1501, DukeMTMC, and MARS respectively.
This verifies the modelling advantages of the 
proposed per-camera tracklet discrimination learning scheme 
on the unsupervised tracklet labels in
inducing cross-view re-id discriminative feature learning.

\begin{table}[h]
	\centering
	\setlength{\tabcolsep}{0.1cm}
	\caption{Effect of Per-Camera Tracklet Discrimination  (PCTD) learning.}
	\label{tab:PCJL}
	\begin{tabular}
		{c||c|c||c|c||c|c||c|c}
		\hline
		Dataset				
		& \multicolumn{2}{c||}{CUHK03\cite{li2014deepreid}}				& \multicolumn{2}{c||}{Market-1501\cite{zheng2015scalable}}				& \multicolumn{2}{c||}{DukeMTMC\cite{ristani2016MTMC}}		& \multicolumn{2}{c}{MARS\cite{zheng2016mars}}			\\ \hline 
		Metric(\%)	& R1	& mAP		& R1	& mAP		& R1	& mAP			& R1	& mAP
		\\ \hline \hline
		JCC		& 29.8	& 12.5		& 17.5	& 7.9		& 14.9	& 3.5			& 18.1	& 13.1
		\\ \hline
		PCTD		& \textbf{33.8}	& \textbf{18.9}		& \textbf{52.1}		& \textbf{26.6}
					& \textbf{51.2}	& \textbf{32.9}		& \textbf{38.0}		& \textbf{23.9}
		\\ \hline
	\end{tabular}
\end{table}

\noindent \textbf{Effectiveness of Cross-Camera Tracklet Association } The CCTA learning component
was evaluated by testing the performance drop after eliminating it.
Table \ref{tab:TA} shows a significant performance benefit
from this model component, 
e.g. a Rank-1 boost of 
10.9\%, 11.6\%, 10.5\%, and 5.8\% 
on CUHK03, Market-1501, DukeMTMC, and MARS respectively.
This validates the importance of modelling the correlation across 
cameras in discriminative optimisation
and the effectiveness of our CCTA deep learning objective loss formulation
in an end-to-end manner.
Additionally, this also suggests the 
effectiveness of the PCTD model component
in facilitating the cross-view identity discrimination learning
by providing re-id sensitive features
in a joint incremental learning manner.

\begin{table}[h]
	\centering
	\setlength{\tabcolsep}{0.1cm}
	\caption{Effect of Cross-Camera Tracklet Association (CCTA)}
	\label{tab:TA}
	\begin{tabular}
		{c||c|c||c|c||c|c||c|c}
		\hline
		Dataset				
		& \multicolumn{2}{c||}{CUHK03\cite{li2014deepreid}}				& \multicolumn{2}{c||}{Market-1501\cite{zheng2015scalable}}				& \multicolumn{2}{c||}{DukeMTMC\cite{zheng2017unlabeled}}		& \multicolumn{2}{c}{MARS\cite{zheng2016mars}}			\\ \hline
		CCTA		& R1	& mAP		& R1	& mAP			& R1	& mAP			& R1	& mAP
		\\ \hline \hline
		\xmark		& 33.8	& 18.9		& 52.1	& 26.6		& 51.2	& 32.9		& 38.0	& 23.9
		\\ \hline
		\cmark		
		& \textbf{44.7}	& \textbf{31.2}		& \textbf{63.7}	& \textbf{41.2}			& \textbf{61.7}	& \textbf{43.5}				& \textbf{43.8}		& \textbf{29.1}
		\\ \hline
	\end{tabular}
\end{table}

\noindent \textbf{Model Robustness Analysis } Finally, we performed an analysis
on model robustness against person ID duplication rates in tracklet labelling.
We conducted a controlled evaluation on MARS 
where multiple tracklets per ID per camera are available for 
setting simulation.
Recall that the ID duplication may mainly 
come with imperfect temporal sampling due to trajectory fragmentation
and when some people stay in the same camera view for a longer
time period than the temporal sampling gap.
To simulate such a situation,
we assume a varying percentage (10\%$\sim$50\%) of IDs per camera have
two random tracklets sampled and annotated with different tracklet labels.
More tracklets per ID per camera are likely to be sampled,
which can make this analysis more complex due to 
the interference from the number of duplicated person IDs.
Table \ref{tab:robustness} shows that 
our TAUDL model is robust against the ID duplication rate,
e.g. with only a Rank-1 drop of 3.1\% given 
as high as 50\% per-camera ID duplication rate.
In reality, it is not too hard to minimise ID duplication rate among tracklets
(Sec. \ref{sec:tracklet_labelling}),
e.g. conducting very sparse sampling over time and space. Note, we do
not care about exhaustive sampling of all the tracklets from video in
a given time period. The model learning benefits from very sparse and
diverse tracklet sampling from a large pool of unlabelled video data.

The robustness of our TAUDL comes with two model components:
{\bf (1)} The model learning optimisation is not only subject to
a single per-camera tracklet label constraint,
but also concurrently to the constraints of all cameras.
This facilitates optimising cross-camera 
tracklet association globally across all cameras in a common
space, due to the Per-Camera Tracklet Discrimination learning mechanism
(Eq. \eqref{eq:PCTD_loss}).
This provides model learning tolerance against
per-camera tracklet label duplication errors.
{\bf (2)} The cross-camera tracklet association learning 
is designed as a feature similarity based ``soft'' objective learning
constraint (Eq. \eqref{eq:CCTA}), without a direct dependence on the
tracklet ID labels.
Therefore, the ID duplication rate has little effect on 
this objective loss constraint.

\begin{table}[h]
	\centering
	\setlength{\tabcolsep}{0.2cm}
	\caption{
		Model robustness analysis on varying ID duplication rates on MARS \cite{zheng2016mars}.
	}
	\label{tab:robustness}
	\begin{tabular}
		{c||c|c|c|c||c}
		\hline
		ID Duplication Rate (\%)	
					& Rank-1 	& Rank-5	& Rank-10	& Rank-20	& mAP  	\\ \hline \hline
		0	 		& \bf 43.8		&\bf  59.9		&\bf  66.0		&\bf  72.8		&\bf 29.1
		\\ \hline
		10	 		& 42.8		& 59.7		& 65.5		& 71.6		& 28.3
		\\ \hline
		20	 		& 42.2		& 58.8		& 64.7		& 70.6		& 27.4
		\\ \hline
		30	 		& 41.6		& 57.9		& 64.5		& 69.7		& 26.7
		\\ \hline
		50 			& 40.7		& 57.0		& 63.4		& 69.6		& 25.6
		\\ \hline
	\end{tabular}
\end{table}

\section{Conclusions}

In this work, we presented a novel {\em Tracklet Association Unsupervised Deep Learning} (TAUDL) model for
unsupervised person re-identification using unsupervised person
tracklet data extracted from videos, therefore eliminating the
tedious and exhaustive manual labelling required by all supervised
learning based re-id model learning. This enables TAUDL to be much
more scalable to real-world re-id deployment at large scale video data. 
In contrast to most existing re-id methods
that either require exhaustively pairwise labelled training data for
every camera pair or 
assume the availability of additional labelled source domain training data for target domain adaptation, 
the proposed TAUDL model is capable of 
end-to-end deep learning a discriminative person re-id model from scratch
on totally unlabelled tracklet data. 
This is achieved by optimising jointly both
the Per-Camera Tracklet Discrimination loss function
and 
the Cross-Camera Tracklet Association loss function
in a single end-to-end deep learning framework.
To our knowledge, this is the first completely unsupervised learning
based re-id model without any identity labels for model learning,
neither pairwise cross-view image pair labelling nor single-view image
identity class labelling.
Extensive comparative evaluations were conducted on six image and video based re-id benchmarks 
to validate the advantages of
the proposed TAUDL model over a wide range of state-of-the-art 
unsupervised and domain adaptation re-id methods.
We also conducted in-depth TAUDL model component evaluation
and robustness test to give insights on 
model performance advantage and model learning stability.

\section*{Acknowledgments}
\small
This work is partially supported by the China Scholarship Council, 
Vision Semantics Limited,
National Natural Science Foundation of China (Project No. 61401212), 
the Key Technology Research and Development Program of Jiangsu Province (Project No. BE2015162), the Science and Technology Support Project of Jiangsu Province (Project No. BE2014714),
Royal Society Newton Advanced Fellowship Programme (NA150459),
and Innovate UK Industrial Challenge Project on Developing and Commercialising Intelligent Video Analytics Solutions for Public Safety (98111-571149).

%
%
%
 \bibliographystyle{splncs04}
 \bibliography{mybibtex}

\begin{thebibliography}{10}
\providecommand{\url}[1]{\texttt{#1}}
\providecommand{\urlprefix}{URL }
\providecommand{\doi}[1]{https://doi.org/#1}

\bibitem{ahmed2015improved}
Ahmed, Jones, Marks: An improved deep learning architecture for person
  re-identification. In: CVPR (2015)

\bibitem{ando2005framework}
Ando, R.K., Zhang, T.: A framework for learning predictive structures from
  multiple tasks and unlabeled data. JMLR  \textbf{6},  1817--1853 (2005)

\bibitem{baltieri20113dpes}
Baltieri, D., Vezzani, R., Cucchiara, R.: 3dpes: 3d people dataset for
  surveillance and forensics. In: J-HGBU (2011)

\bibitem{bautista2017deep}
Bautista, M.A., Sanakoyeu, A., Ommer, B.: Deep unsupervised similarity learning
  using partially ordered sets. In: CVPR (2017)

\bibitem{bautista2016cliquecnn}
Bautista, M.A., Sanakoyeu, A., Tikhoncheva, E., Ommer, B.: Cliquecnn: Deep
  unsupervised exemplar learning. In: NIPS (2016)

\bibitem{chen2018group}
Chen, D., Xu, D., Li, H., Sebe, N., Wang, X.: Group consistent similarity
  learning via deep crf for person re-identification. In: CVPR (2018)

\bibitem{chen2017beyond}
Chen, W., Chen, X., Zhang, J., Huang, K.: Beyond triplet loss: a deep
  quadruplet network for person re-identification. In: CVPR (2017)

\bibitem{chen2017person}
Chen, Y., Zhu, X., Gong, S.: Person re-identification by deep learning
  multi-scale representations. In: ICCV Workshop (2017)

\bibitem{chen2018person}
Chen, Y.C., Zhu, X., Zheng, W.S., Lai, J.H.: Person re-identification by camera
  correlation aware feature augmentation. IEEE TPAMI  \textbf{40}(2),  392--408
  (2018)

\bibitem{cheng2016person}
Cheng, D., Gong, Y., Zhou, S., Wang, J., Zheng, N.: Person re-identification by
  multi-channel parts-based cnn with improved triplet loss function. In: CVPR
  (2016)

\bibitem{cho2016improving}
Cho, Y.J., Yoon, K.J.: Improving person re-identification via pose-aware
  multi-shot matching. In: CVPR (2016)

\bibitem{evgeniou2004regularized}
Evgeniou, T., Pontil, M.: Regularized multi--task learning. In: SIGKDD (2004)

\bibitem{fan2017unsupervised}
Fan, H., Zheng, L., Yang, Y.: Unsupervised person re-identification: Clustering
  and fine-tuning. arXiv preprint arXiv:1705.10444  (2017)

\bibitem{farenzena2010person}
Farenzena, M., Bazzani, L., Perina, A., Murino, V., Cristani, M.: Person
  re-identification by symmetry-driven accumulation of local features. In: CVPR
  (2010)

\bibitem{gong2014person}
Gong, S., Cristani, M., Yan, S., Loy, C.C.: Person re-identification. Springer
  (2014)

\bibitem{gray2008viewpoint}
Gray, D., Tao, H.: Viewpoint invariant pedestrian recognition with an ensemble
  of localized features. In: ECCV (2008)

\bibitem{he2016deep}
He, K., Zhang, X., Ren, S., Sun, J.: Deep residual learning for image
  recognition. In: CVPR (2016)

\bibitem{hermans2017defense}
Hermans, A., Beyer, L., Leibe, B.: In defense of the triplet loss for person
  re-identification. arXiv preprint arXiv:1703.07737  (2017)

\bibitem{hirzer2011person}
Hirzer, M., Beleznai, C., Roth, P.M., Bischof, H.: Person re-identification by
  descriptive and discriminative classification. In: SCIA (2011)

\bibitem{jiao2018deep}
Jiao, J., Zheng, W.S., Wu, A., Zhu, X., Gong, S.: Deep low-resolution person
  re-identification. In: AAAI (2018)

\bibitem{kawanishi2014shinpuhkan2014}
Kawanishi, Y., Wu, Y., Mukunoki, M., Minoh, M.: Shinpuhkan2014: A multi-camera
  pedestrian dataset for tracking people across multiple cameras. In: FCV
  (2014)

\bibitem{khan2016unsupervised}
Khan, F.M., Bremond, F.: Unsupervised data association for metric learning in
  the context of multi-shot person re-identification. In: AVSS (2016)

\bibitem{kingma2014adam}
Kingma, D.P., Ba, J.: Adam: A method for stochastic optimization. arXiv
  preprint arXiv:1412.6980  (2014)

\bibitem{kodirov2016person}
Kodirov, E., Xiang, T., Fu, Z., Gong, S.: Person re-identification by
  unsupervised $l_1$ graph learning. In: ECCV (2016)

\bibitem{kodirov2015dictionary}
Kodirov, E., Xiang, T., Gong, S.: Dictionary learning with iterative laplacian
  regularisation for unsupervised person re-identification. In: BMVC (2015)

\bibitem{leal2015motchallenge}
Leal-Taix{\'e}, L., Milan, A., Reid, I., Roth, S., Schindler, K.: Motchallenge
  2015: Towards a benchmark for multi-target tracking. arXiv preprint
  arXiv:1504.01942  (2015)

\bibitem{lee2008sparse}
Lee, H., Ekanadham, C., Ng, A.Y.: Sparse deep belief net model for visual area
  v2. In: NIPS (2008)

\bibitem{li2012human}
Li, W., Zhao, R., Wang, X.: Human reidentification with transferred metric
  learning. In: ACCV (2012)

\bibitem{li2014deepreid}
Li, W., Zhao, R., Xiao, T., Wang, X.: Deepreid: Deep filter pairing neural
  network for person re-identification. In: CVPR (2014)

\bibitem{li2017person}
Li, W., Zhu, X., Gong, S.: Person re-identification by deep joint learning of
  multi-loss classification. In: IJCAI (2017)

\bibitem{li2018harmonious}
Li, W., Zhu, X., Gong, S.: Harmonious attention network for person
  re-identification. In: CVPR (2018)

\bibitem{lisanti2015person}
Lisanti, G., Masi, I., Bagdanov, A.D., Del~Bimbo, A.: Person re-identification
  by iterative re-weighted sparse ranking. IEEE TPAMI  \textbf{37}(8),
  1629--1642 (2015)

\bibitem{liu2014semi}
Liu, X., Song, M., Tao, D., Zhou, X., Chen, C., Bu, J.: Semi-supervised coupled
  dictionary learning for person re-identification. In: CVPR (2014)

\bibitem{liu2017quality}
Liu, Y., Yan, J., Ouyang, W.: Quality aware network for set to set recognition.
  In: CVPR (2017)

\bibitem{liu2017stepwise}
Liu, Z., Wang, D., Lu, H.: Stepwise metric promotion for unsupervised video
  person re-identification. In: ICCV (2017)

\bibitem{loy_ijcv2010}
Loy, C., Xiang, T., Gong, S.: Time-delayed correlation analysis for
  multi-camera activity understanding. IJCV  \textbf{90}(1),  106--129 (2010)

\bibitem{ma2017person}
Ma, X., Zhu, X., Gong, S., Xie, X., Hu, J., Lam, K.M., Zhong, Y.: Person
  re-identification by unsupervised video matching. Pattern Recognition
  \textbf{65},  197--210 (2017)

\bibitem{peng2016unsupervised}
Peng, P., Xiang, T., Wang, Y., Pontil, M., Gong, S., Huang, T., Tian, Y.:
  Unsupervised cross-dataset transfer learning for person re-identification.
  In: CVPR (2016)

\bibitem{prosser2010person}
Prosser, B.J., Zheng, W.S., Gong, S., Xiang, T.: Person re-identification by
  support vector ranking. In: BMVC (2010)

\bibitem{qin2015unsupervised}
Qin, C., Song, S., Huang, G., Zhu, L.: Unsupervised neighborhood component
  analysis for clustering. Neurocomputing  \textbf{168},  609--617 (2015)

\bibitem{ristani2016MTMC}
Ristani, E., Solera, F., Zou, R., Cucchiara, R., Tomasi, C.: Performance
  measures and a data set for multi-target, multi-camera tracking. In: ECCV
  Workshop (2016)

\bibitem{ristani2016performance}
Ristani, E., Solera, F., Zou, R., Cucchiara, R., Tomasi, C.: Performance
  measures and a data set for multi-target, multi-camera tracking. In: ECCV
  Workshop (2016)

\bibitem{su2016deep}
Su, C., Zhang, S., Xing, J., Gao, W., Tian, Q.: Deep attributes driven
  multi-camera person re-identification. In: ECCV (2016)

\bibitem{subramaniam2016deep}
Subramaniam, Chatterjee, Mittal: Deep neural networks with inexact matching for
  person re-identification. In: NIPS (2016)

\bibitem{Ustinova2016hist}
Ustinova, E., Lempitsky, V.: Learning deep embeddings with histogram loss. In:
  NIPS (2016)

\bibitem{wang2016joint}
Wang, F., Zuo, W., Lin, L., Zhang, D., Zhang, L.: Joint learning of
  single-image and cross-image representations for person re-identification.
  In: CVPR (2016)

\bibitem{wang2014unsupervised}
Wang, H., Gong, S., Xiang, T.: Unsupervised learning of generative topic
  saliency for person re-identification. In: BMVC (2014)

\bibitem{wang2018person}
Wang, H., Zhu, X., Gong, S., Xiang, T.: Person re-identification in identity
  regression space. IJCV  (2018)

\bibitem{wang2016towards}
Wang, H., Zhu, X., Xiang, T., Gong, S.: Towards unsupervised open-set person
  re-identification. In: ICIP (2016)

\bibitem{want2018Transfer}
Wang, J., Zhu, X., Gong, S., Li, W.: Transferable joint attribute-identity deep
  learning for unsupervised person re-identification. In: CVPR (2018)

\bibitem{wang2014person}
Wang, T., Gong, S., Zhu, X., Wang, S.: Person re-identification by video
  ranking. In: ECCV (2014)

\bibitem{wang2016person}
Wang, T., Gong, S., Zhu, X., Wang, S.: Person re-identification by
  discriminative selection in video ranking. IEEE TPAMI  \textbf{38}(12),
  2501--2514 (2016)

\bibitem{xiao2016learning}
Xiao, T., Li, H., Ouyang, W., Wang, X.: Learning deep feature representations
  with domain guided dropout for person re-identification. In: CVPR (2016)

\bibitem{ye2007adaptive}
Ye, J., Zhao, Z., Liu, H.: Adaptive distance metric learning for clustering.
  In: CVPR (2007)

\bibitem{ye2017dynamic}
Ye, M., Ma, A.J., Zheng, L., Li, J., Yuen, P.C.: Dynamic label graph matching
  for unsupervised video re-identification. In: ICCV (2017)

\bibitem{yu2017cross}
Yu, H.X., Wu, A., Zheng, W.S.: Cross-view asymmetric metric learning for
  unsupervised person re-identification. In: ICCV (2017)

\bibitem{zhang2016far}
Zhang, S., Benenson, R., Omran, M., Hosang, J., Schiele, B.: How far are we
  from solving pedestrian detection? In: CVPR (2016)

\bibitem{zhang2017deep}
Zhang, Y., Xiang, T., Hospedales, T.M., Lu, H.: Deep mutual learning. In: CVPR
  (2018)

\bibitem{zhao2017person}
Zhao, R., Oyang, W., Wang, X.: Person re-identification by saliency learning.
  IEEE TPAMI  \textbf{39}(2),  356--370 (2017)

\bibitem{zheng2016mars}
Zheng, L., Bie, Z., Sun, Y., Wang, J., Su, C., Wang, S., Tian, Q.: Mars: A
  video benchmark for large-scale person re-identification. In: ECCV (2016)

\bibitem{zheng2015scalable}
Zheng, L., Shen, L., Tian, L., Wang, S., Wang, J., Tian, Q.: Scalable person
  re-identification: A benchmark. In: CVPR (2015)

\bibitem{zheng2017unlabeled}
Zheng, Z., Zheng, L., Yang, Y.: Unlabeled samples generated by gan improve the
  person re-identification baseline in vitro. In: ICCV (2017)

\bibitem{zhu2017fast}
Zhu, X., Wu, B., Huang, D., Zheng, W.S.: Fast openworld person
  re-identification. IEEE TIP pp. 2286--2300 (2017)

\end{thebibliography}

%
%
%
%
%
\end{document}